\documentclass[12pt,a4paper]{article}

\usepackage[british]{babel}

\usepackage[a4paper,top=2cm,bottom=2cm,left=2.5cm,right=2.5cm,marginparwidth=1.75cm]{geometry}

\usepackage[style=apa, backend=biber]{biblatex} % APA 7th edition style citations using biblatex
\DeclareLanguageMapping{british}{british-apa} % Set language mapping
\DeclareFieldFormat[article]{volume}{\apanum{#1}} % Format volume number

\usepackage{amsmath}
\usepackage{graphicx}
\usepackage[colorlinks=true, allcolors=blue]{hyperref}
\usepackage{hyperref}
\usepackage[title]{appendix}
\usepackage{mathrsfs}
\usepackage{amsfonts}
\usepackage{booktabs} % For \toprule, \midrule, \botrule
\usepackage{caption}  % For \caption
\usepackage{threeparttable} % For table footnotes
\usepackage{algorithm}
\usepackage{algorithmicx}
\usepackage{algpseudocode}
\usepackage{listings}
\usepackage{enumitem}
\usepackage{chngcntr}
\usepackage{booktabs}
\usepackage{lipsum}
\usepackage{subcaption}
\usepackage{authblk}
\usepackage[T1]{fontenc}    % Font encoding
\usepackage{csquotes}       % Include csquotes
\usepackage{diagbox}
\usepackage{comment}

\usepackage{subcaption}
\usepackage[justification=raggedright, singlelinecheck=false]{caption}
\usepackage{caption}
\usepackage{helvet}  % Load Helvetica (Arial-like)
\usepackage{lmodern}  % Ensures default LaTeX font is available

\usepackage{setspace}
\usepackage{titlesec}
\titleformat{\section} % Redefine section numbering format
  {\normalfont\Large\bfseries}{\thesection.}{1em}{}
  
\usepackage{float}   % for customizing caption position
\usepackage{caption} % for customizing caption format


\title{Cross-Modal Deep Metric Learning for Time Series Anomaly Detection}

\author[1]{Wei Li}
\author[1]{Zheze Yang}

\affil[1]{\small The University of Texas at Austin, Austin, USA}
\affil[*]{\small Corresponding author. \texttt{zezey745@gmail.com}}

\date{}  % Remove date

\begin{document}
\maketitle
\begin{abstract}
To effectively address the issues of low sensitivity and high time consumption in time series anomaly detection, we propose an anomaly detection method based on cross-modal deep metric learning. A cross-modal deep metric learning feature clustering model is constructed, composed of an input layer, a triplet selection layer, and a loss function computation layer. The squared Euclidean distances between cluster centers are calculated, and a stochastic gradient descent strategy is employed to optimize the model and classify different time series features. The inner product of principal component direction vectors is used as a metric for anomaly measurement. The von Mises–Fisher (vMF) distribution is applied to describe the directional characteristics of time series data, and historical data is used to train and obtain evaluation parameters. By comparing the principal component direction vector of actual time series data with the threshold, anomaly detection is performed. Experimental results demonstrate that the proposed method accurately classifies time series data with different attributes, exhibits high sensitivity to anomalies, and achieves high detection accuracy, fast detection speed, and strong robustness.

\end{abstract}


\section{Introduction}

In recent years, deep learning, as an important branch of the field of machine learning, has attracted much attention in various domains. It has provided new perspectives for the study and application of deep learning theories and methods. Common deep learning models include Convolutional Neural Networks (CNN), Recurrent Neural Networks (RNN) with “memory” capability, Long Short-Term Memory (LSTM), Deep Belief Networks (DBN), and Restricted Boltzmann Machines (RBM), among others.

Financial risk measurement is a fundamental task in predicting financial market fluctuations. Among them, Value at Risk (VaR) is one of the most widely used standards in risk management [1]. VaR quantifies the maximum potential loss of a financial asset or portfolio within a given confidence level over a specific time horizon. Methods for calculating return volatility can be categorized into three types: (1) parametric methods, such as assuming normal or t-distributions for returns; these approaches are simple but often deviate from reality; (2) non-parametric methods, such as historical simulation, which directly use past return data without distributional assumptions; and (3) semi-parametric methods, such as Extreme Value Theory (EVT), which focus on the tail behavior of return distributions. To address the limitations of traditional volatility measures, realized volatility (RV) based on high-frequency data was later introduced[2], and long-memory properties of RV were subsequently modeled through the Heterogeneous Autoregressive (HAR) framework [3].
Volatility prediction plays a central role in VaR risk management and continues to be a major topic in financial research. The concept of realized volatility (RV) based on high-frequency data has been proposed [4], offering a more accurate reflection of market fluctuations than traditional volatility measures. Subsequent studies have shown that RV series display long-memory characteristics, motivating the use of the ARFIMA model to capture RV dynamics and improve forecasting performance [5]. To further model these features, the Heterogeneous Autoregressive (HAR) framework was developed [6], and later extended to variants such as HARQ and HARF to enhance predictive capability [7]. Empirical evidence consistently demonstrates that approaches relying on high-frequency data outperform conventional low-frequency GARCH-type models in volatility prediction.

With the rapid development of deep learning, its adoption in financial research has grown significantly. Neural networks have been applied to volatility forecasting, where prediction metrics incorporating frequency domain characteristics have been introduced [8][9]. In addition, LSTM models have been employed in combination with stock trading data for volatility prediction, and their performance has been compared with traditional approaches such as GARCH, ARFIMA, and HAR. Results suggest that LSTM-based methods can achieve superior predictive accuracy under certain conditions.

This study uses trading data combined with deep learning LSTM models to apply VaR measurement methods to financial risk management. In addition, the paper will discuss how to use financial market price data to construct a realized volatility sequence and a realized return sequence, then use LSTM modeling to predict RV, and integrate it with EVT to form an LSTM–RV–EVT financial risk management VaR model for empirical analysis.

\section{Related work}
Deep learning has become a cornerstone for time series analysis, offering powerful tools to model complex temporal dynamics that traditional statistical models often fail to capture. Foundational work in this area has heavily relied on recurrent architectures, such as Long Short-Term Memory (LSTM) and Gated Recurrent Unit (GRU) networks, which are specifically designed to handle long-range dependencies. Hybrid LSTM-GRU architectures, for instance, have been effectively utilized for temporal dependency modeling in financial applications like loan default prediction [10]. The capabilities of these models are further extended in forecasting corporate financial metrics, where LSTMs have demonstrated significant improvements over conventional methods [11]. To better capture intricate patterns, hybrid models combining LSTMs with other architectures, such as Convolutional Neural Networks (CNNs) and Transformers, have been developed for tasks like financial volatility forecasting, showcasing the power of architectural synergy [12]. Beyond standard recurrence, bidirectional LSTMs integrated with multi-scale attention mechanisms provide a more robust framework for sequence mining by processing data in both forward and backward directions, thereby enriching the contextual representation [13].

The Transformer architecture, originally developed for natural language processing, has also proven to be a paradigm-shifting innovation for time series analysis due to its superior ability to model temporal dependencies via self-attention mechanisms. Its application in stock price prediction has highlighted its strength in capturing both long-term trends and short-term fluctuations from multi-dimensional feature sets [14]. The principles of attention and temporal modeling from Transformers have been adapted for vision-oriented tasks like multi-object tracking, demonstrating the cross-domain applicability of these core concepts for sequential data processing [15]. Further, hybrid approaches that combine LSTM models with statistical methods like the copula function have been proposed to forecast risk in multi-asset portfolios, effectively modeling the complex and nonlinear dependencies between different time series [16]. A deep learning framework that integrates CNN and BiLSTM has also been proposed for systemic risk analysis [17]. These advancements in sequential modeling provide a strong methodological basis for our work in extracting discriminative features from time series data for anomaly detection.
Beyond linear sequences, many time series anomalies are embedded within complex relational structures. To address this, graph-based models have emerged as a powerful methodology for capturing intricate dependencies. Temporal graph representation learning, for instance, has been used to model the evolving behavior of users in transactional networks, providing a dynamic structural perspective on anomaly detection [18]. The synergy between graph-based and sequential models is evident in frameworks that jointly apply graph convolution and recurrent modeling, enabling scalable and accurate estimation in domains like network traffic analysis [19]. This integration of architectures, such as combining Transformers with transaction graphs for monitoring financial risks, allows for the simultaneous capture of both relational and temporal patterns [20]. The effectiveness of these graph neural networks (GNNs) is often enhanced through entity-aware modeling to improve structured information extraction in specific domains [21] or through attention mechanisms optimized to identify malicious user patterns in complex systems [22].

Central to anomaly detection is the task of learning discriminative data representations, a challenge addressed by various representation learning paradigms. Unsupervised methods based on contrastive learning are particularly relevant, as they excel in settings like fraud detection where labeled data is scarce by learning to differentiate between normal and anomalous patterns [23]. The performance of contrastive learning itself can be enhanced through a systematic analysis of data augmentation techniques, which helps in creating more robust and generalizable representations [24]. Alternatively, deep generative models offer another approach by learning the underlying distribution of normal data, thereby identifying anomalies as low-probability events in complex financial transactions [25]. More advanced techniques in causal representation learning aim to build robust models by uncovering and modeling the underlying causal factors in the data generation process [26], with some methods focusing on target-oriented learning for improved cross-domain prediction [27]. These representation learning principles directly inform our approach to deep metric learning. Furthermore, leveraging information from multiple sources or modalities can significantly enhance detection accuracy. Methodologies for multi-modal detection that incorporate feature alignment and modality-specific attention have been developed [28], while multi-task learning frameworks have been used to fuse cross-domain data for macroeconomic forecasting [29]. The fusion of features from different modalities, such as in CNN-Transformer approaches, has also shown success in classification tasks [30].
Reinforcement learning (RL) provides a distinct paradigm for anomaly detection and risk management, framing the problem as one of optimal sequential decision-making. Methodologies in this area include nested RL strategies for dynamic risk control in financial markets [31] and the use of Deep Q-Networks (DQN) for optimizing tasks in dynamic environments like operating system scheduling [32] and backend cache management [33]. RL has also been applied to enable autonomous resource management in microservice systems [34] and for adaptive container migration in cloud-native environments [35]. In multi-agent settings, collaborative RL approaches have been developed for tasks such as elastic cloud resource scaling [36] and adaptive resource orchestration [37]. Further applications in finance have explored QTRAN-based frameworks for portfolio optimization [38].

\section{Long Short-Term Memory (LSTM) Model}

The standard recurrent neural network (RNN) is a type of short-term memory neural network. In the structure of recurrent neural networks, long-term dependence problems may occur during the training process, leading to gradient vanishing or exploding issues. In practical applications, researchers have introduced variants of RNNs, such as Long Short-Term Memory (LSTM) and Gated Recurrent Unit (GRU), with LSTM generally producing better results. This paper applies the LSTM network to VaR risk measurement of financial assets. Compared with traditional RNN models, the LSTM network adds a memory cell to store information and introduces a gating mechanism to control the amount of information flowing into and out of the cell. This allows the network to selectively remember or forget past information, effectively addressing the gradient vanishing problem of traditional RNNs.

The LSTM memory cell includes three gates: input gate, forget gate, and output gate. Each gate uses a sigmoid activation function with output values between 0 and 1, determining how much information passes through. The forget gate controls the memory retention of past cell states, enabling selective forgetting of historical information. The formulas are defined as follows:

\[
i_t = \sigma(W_{xi} x_t + W_{hi} h_{t-1} + b_i)
\]
\[
f_t = \sigma(W_{xf} x_t + W_{hf} h_{t-1} + b_f)
\]
\[
\tilde{c}_t = \tanh(W_{xc} x_t + W_{hc} h_{t-1} + b_c)
\]
\[
c_t = f_t \cdot c_{t-1} + i_t \cdot \tilde{c}_t
\]
\[
o_t = \sigma(W_{xo} x_t + W_{ho} h_{t-1} + b_o)
\]
\[
h_t = o_t \cdot \tanh(c_t)
\]

Here, $x_t$ is the input at time $t$, $h_t$ is the output of the LSTM cell, and $c_t$ is the cell state. The gates $i_t$, $f_t$, and $o_t$ control the update and output of the cell state, enabling the LSTM to capture long-term dependencies in the sequence.

\section{Model Construction}

Value at Risk (VaR) refers to the maximum potential loss of a financial asset or portfolio within a specified time period at a given confidence level. The normalized return of a financial asset at time $t$ can be expressed as:
\[
r_t = \frac{R_t - \mu_t}{\sigma_t}
\]
where $n_t$ denotes the normalized return, $R_t$ is the asset return, $\mu_t$ is the mean, and $\sigma_t$ is the standard deviation of returns. The VaR under confidence level $(1-p_0)$ is given by:
\[
VaR_t = F^{-1}(1-p_0) \sigma_{t+1}
\]

Here, $F^{-1}$ denotes the inverse cumulative distribution function, and $\sigma_{t+1}$ is the predicted standard deviation for day $t+1$ based on data up to day $t$.

\subsection{Realized Volatility}

Andersen and Bollerslev (1998) proposed realized volatility (RV) as a measure of volatility using high-frequency intraday returns:
\[
RV(p_t) = \sum_{i=1}^{N(\Delta)} (\ln P_{t,i} - \ln P_{t,i-1})^2
\]
where $P_{t,i}$ is the price at the $i$-th intraday time point on day $t$, and $N(\Delta)$ is the number of intraday observations per day.

\subsection{Wavelet-LSTM-RV Dynamic Volatility Prediction Model}

Since volatility series often exhibit nonstationarity and long memory, LSTM is suitable for modeling such time series. This paper uses transaction prices and volumes to construct the LSTM–RV prediction model as follows:
\[
\ln RV(P_t) = LSTM(\ln RV(P_{t-1}), \dots, \ln RV(P_{t-p}))
\]
\[
\ln RV(V_t) = LSTM(\ln RV(V_{t-1}), \dots, \ln RV(V_{t-p}))
\]

The realized volatility sequence $RV(P_t)$ is obtained from daily transaction price data, and the realized volatility sequence $RV(V_t)$ is obtained from daily transaction volume data. The input variables are the univariate $RV$ series, which are processed using a univariate LSTM model. This approach has two advantages: (1) For price-based RV, it uses historical statistical results and prediction errors; (2) For volume-based RV, it uses trading volume data to predict volatility, thus incorporating both price and volume information to improve volatility prediction accuracy.

\subsection{LSTM–RV–EVT Model Construction}
From equation (2), it is apparent that the estimation of Value at Risk (VaR) is governed by both the predicted volatility and the quantiles of the return distribution. During periods of heightened market turbulence, the return distribution is frequently observed to possess fat tails and significant kurtosis, which renders the conventional normality assumption unsuitable for accurate risk assessment.

To address these challenges, this study adopts a semi-parametric quantile estimation framework based on Extreme Value Theory (EVT). The selection and integration of EVT into the methodology are directly informed by the insights of Yan et al.[39], who emphasized the necessity of moving beyond parametric assumptions in the modeling of financial extremes. Their work provided a technical foundation for employing distributional models, such as the generalized Pareto distribution (GPD), to effectively capture the statistical characteristics of tail risk. Specifically, they also demonstrated that supplementing deep learning models with EVT-based tail modeling improves the accuracy and robustness of risk estimation under non-Gaussian conditions---a finding that directly informs the structure of this study's approach. Motivated by their results, the present methodology incorporates the GPD to model the distributional tail, thereby allowing for precise quantile estimation in extreme value regions. The cumulative distribution function (CDF) of the GPD, as detailed in their work, is utilized here to support reliable VaR calculation as follows:

% From equation (2), we know that VaR is determined by the predicted volatility and return distribution quantiles. In cases of extreme fluctuations, the return distribution often shows fat tails and leptokurtosis, making it inappropriate to use a normal distribution assumption. This paper adopts a semi-parametric approach, estimating the quantiles of the return distribution using Extreme Value Theory (EVT). EVT assumes that the distribution tail follows a generalized Pareto distribution (GPD), with its cumulative distribution function (CDF) given by:
\[
F_{\xi, \beta}(x) =
\begin{cases}
1 - \left(1 + \frac{\xi x}{\beta}\right)^{-1/\xi}, & \xi \neq 0, \\
1 - \exp\left(-\frac{x}{\beta}\right), & \xi = 0,
\end{cases}
\]
where $\xi$ is the shape parameter and $\beta$ is the scale parameter. When $\xi \ge 0$, the GPD has a heavy tail; when $\xi < 0$, it has a bounded tail. Therefore, the $(1-p_0)$ quantile of the distribution can be expressed as:
\[
F^{-1}(1-p_0) = u + \frac{\beta}{\xi} \left[ \left( \frac{p_0}{n_u/n} \right)^{-\xi} - 1 \right],
\]
(4)
where $n$ is the sample size, $n_u$ is the number of exceedances over threshold $u$. Following Neftci (2000), this paper sets the threshold at 1.65 times the standard deviation of returns.

Combining the LSTM model from equation (3) with the EVT model in equation (4), and based on the predicted volatility for the next $t+1$ day, we construct the LSTM–RV–EVT VaR model:
\[
VaR_{t+1|t}^P = F^{-1}(1-p_0) \sqrt{RV(P)_{t+1|t}},
\]
\[
VaR_{t+1|t}^V = F^{-1}(1-p_0) \sqrt{RV(V)_{t+1|t}}.
\]
This hybrid model has the advantages of using high-frequency data for RV estimation, improving volatility prediction accuracy with LSTM, and accounting for heavy-tailed behavior with EVT.

\subsection{ VaR Backtesting}

After obtaining the VaR estimates, we need to backtest them to verify their accuracy. Define the loss indicator function:
\[
I_t =
\begin{cases}
1, & R_t < -VaR_t, \\
0, & R_t \ge -VaR_t,
\end{cases}
\]
where $R_t$ is the actual return. If $I_t=1$, a loss exceeding the VaR occurred on day $t$.

The unconditional coverage (UC) test, independence (IND) test, and conditional coverage (CC) test are used to evaluate the accuracy of VaR models. The UC test examines whether the proportion of violations matches the expected value; the IND test checks whether violations are independent over time; the CC test jointly considers both frequency and independence. Following Candelon et al. (2011), the LR test statistic is:
\[
J_{cc}(l) \sim \chi^2(2),
\]
where $\chi^2(2)$ is the chi-square distribution with 2 degrees of freedom.

\section{Empirical Analysis}

This paper uses high-frequency 5-minute transaction data of the CSI 300 Index from January 4, 2000, to October 31, 2017, totaling 4,239 trading days. The training set covers January 4, 2000, to December 31, 2009, the validation set covers January 4, 2010, to December 31, 2013, and the test set covers January 4, 2014, to October 31, 2017.

We construct the LSTM model using 90\% of the training set for training and 10\% for validation, using early stopping to prevent overfitting. After obtaining the predicted volatility for the next day, we apply the LSTM–RV model to generate VaR forecasts for both price and volume data.
After 10, 20 periods, the Ljung–Box test statistics $Q(5)$, $Q(10)$, and $Q(20)$ are significant at the 1\% level, indicating that the series has autocorrelation, demonstrating its long-memory property.

\begin{table}[H]
\centering
\caption{Descriptive statistics of each constructed series}
\begin{tabular}{lccccc}
\toprule
Statistic & RV(p) & RV(V) & $\ln RV(p)$ & $\ln RV(V)$ & Standardized returns \\
\midrule
Mean         & 0.0006 & 0.3498 & -9.2708 & -1.0439 & -0.1110 \\
Median       & 0.0003 & 0.3464 & -8.9904 & -1.0612 & -0.0002 \\
Std. Dev.    & 5.6860 & 0.0928 & 0.5410 & 0.1432 & 1.0000 \\
Skewness     & 80.5152 & 5.3804 & -0.7346 & 0.6033 & -0.3528 \\
Kurtosis     & 6480.7 & 71.1222 & 7.9966 & 5.7272 & 7.0313 \\
J–B          & 4807207.8$^{**}$ & 172153.3$^{**}$ & 7126.06$^{**}$ & 1266.92$^{**}$ & 1922.96$^{**}$ \\
Q(5)         & 9785$^{**}$ & 4710.8$^{**}$ & 7972.2$^{**}$ & 4202.9$^{**}$ & 39.5$^{**}$ \\
Q(10)        & 15862$^{**}$ & 7402.5$^{**}$ & 12768$^{**}$ & 6595.6$^{**}$ & 56.9$^{**}$ \\
Q(20)        & 15462$^{**}$ & 12092$^{**}$ & 21633$^{**}$ & 10983$^{**}$ & 75.7$^{**}$ \\
\bottomrule
\end{tabular}
\begin{flushleft}
\footnotesize{$^{**}$ denotes significance at the 1\% level. $Q(s)$ is the Ljung–Box statistic for lag $s$.}
\end{flushleft}
\end{table}

\subsection{Volatility Prediction Model Comparison}

This paper compares six volatility prediction models using high-frequency data: LSTM–RV, LSTM, HAR, HAR–QF, and ARFIMA. The LSTM–RV model is based on equation (3) and uses high-frequency price data and trading volume data. LSTM refers to the model using only high-frequency price data. The HAR model (Heterogeneous Autoregressive model) proposed by Corsi (2009) uses lagged RV over 1, 5, and 22 days to predict the next day’s RV. Bollerslev et al. (2016) extend the HAR model with realized quarticity, forming the HARQ model. If realized quarticity is less than RV, the HARQ model becomes the HAR model. The ARFIMA model uses log RV as the dependent variable and is selected using the AIC criterion.

We evaluate prediction accuracy by comparing the predicted and actual volatility values, using four common evaluation metrics: mean squared error (MSE), mean absolute error (MAE), mean absolute percentage error (MAPE), and quasi-likelihood loss function (QLIKE). They are defined as:
\[
MSE = \frac{1}{N} \sum_{n=1}^{N} (y_n - \hat{y}_n)^2, \quad
MAE = \frac{1}{N} \sum_{n=1}^{N} |y_n - \hat{y}_n|,
\]
\[
QLIKE = \frac{1}{N} \sum_{n=1}^{N} \left( \frac{y_n}{\hat{y}_n} - \ln \frac{y_n}{\hat{y}_n} - 1 \right), \quad
MAPE = \frac{1}{N} \sum_{n=1}^{N} \left| \frac{y_n - \hat{y}_n}{y_n} \right|.
\]
Here, $y_n$ is the actual value and $\hat{y}_n$ is the predicted value.

Table 2 presents the accuracy metrics for the six volatility prediction models across the four evaluation criteria. From Table 2, we see that the LSTM–RV model outperforms the others, achieving the smallest errors for all four evaluation metrics, with the largest improvement margin. For instance, compared to the HAR model, the LSTM–RV model reduces MSE, MAE, QLIKE, and MAPE by 15.16\%, 23.79\%, 20.96\%, and 21.49\%, respectively. These results show that:
(1) Based on deep learning, the LSTM–RV model has better prediction performance than the HAR and ARFIMA models in volatility forecasting.
(2) The improvements in prediction accuracy demonstrate that incorporating both price and volume data into the LSTM–RV model significantly enhances its forecasting ability, thanks to its long-memory learning capability.

\begin{table}[H]
\centering
\caption{Comparison of prediction accuracy for six volatility models}
\begin{tabular}{lcccc}
\toprule
Model & MSE & MAE & QLIKE & MAPE \\
\midrule
LSTM–RV        & 4.247 & 0.660 & 0.172 & 0.081 \\
LSTM           & 4.448 & 0.712 & 0.185 & 0.085 \\
HAR            & 4.938 & 0.865 & 0.218 & 0.104 \\
HARQ           & 4.993 & 0.889 & 0.229 & 0.106 \\
HARQF          & 4.947 & 0.882 & 0.225 & 0.105 \\
ARFIMA         & 4.956 & 0.892 & 0.229 & 0.109 \\
\bottomrule
\end{tabular}
\end{table}

\begin{figure}[H]
\centering
\includegraphics[width=0.7\linewidth]{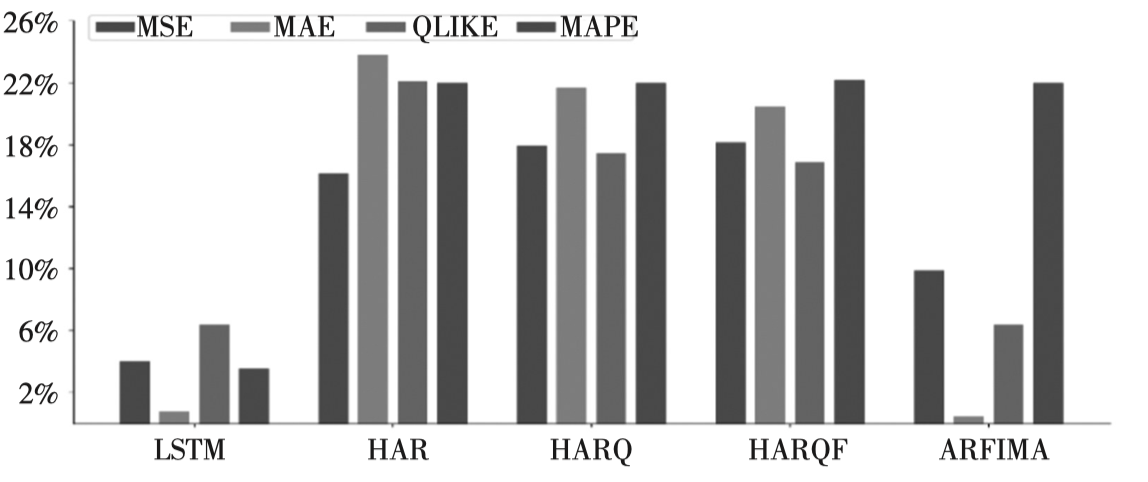}
\caption{Percentage improvement in prediction accuracy of LSTM–RV over five other volatility prediction models}
\end{figure}

\subsection{VaR Backtesting Results}

Table 3 presents the VaR backtesting results for 18 models. The first column lists the VaR model names; the second and third columns are the violation ratios for long and short positions, respectively; the fourth and fifth columns are the UC test statistics $J_c(q)$ and rankings; the sixth and seventh columns are the violation ratios for long and short positions under the corresponding models; the eighth and ninth columns are the CC test statistics and rankings.

The first to sixth rows correspond to models estimated using EVT for quantile calculation; the seventh to fourteenth rows use the skewed Student-$t$ distribution (SKST) for quantile estimation; the fifteenth to eighteenth rows use the historical simulation (H) method.

\begin{table}[H]
\centering
\caption{VaR backtesting results for long and short positions}
\label{tab:var_results}
\begin{tabular}{lcccccccc}
\toprule
Model & \multicolumn{2}{c}{Long positions} & Rank & \multicolumn{2}{c}{Short positions} & Rank \\
 & Violation ratio & $J_c(5)$ & & Violation ratio & $J_c(5)$ & \\
\midrule
LSTM–RV–EVT    & 0.0144 & 0.4601 & 1 & 0.0148 & 0.5949 & 1 \\
LSTM–EVT       & 0.0144 & 0.5335 & 2 & 0.0148 & 0.7954 & 2 \\
HAR–EVT        & 0.0182 & 0.8267 & 6 & 0.0197 & 1.0721 & 6 \\
HARQ–EVT       & 0.0188 & 0.8819 & 7 & 0.0198 & 1.1034 & 7 \\
HARQF–EVT      & 0.0187 & 0.8724 & 5 & 0.0196 & 1.0548 & 5 \\
ARFIMA–EVT     & 0.0201 & 1.0027 & 8 & 0.0214 & 1.1984 & 8 \\
LSTM–RV–SKST   & 0.0145 & 0.4831 & 3 & 0.0149 & 0.6483 & 3 \\
LSTM–SKST      & 0.0145 & 0.5498 & 4 & 0.0149 & 0.8337 & 4 \\
HAR–SKST       & 0.0193 & 0.8685 & 9 & 0.0207 & 1.0803 & 9 \\
HARQ–SKST      & 0.0201 & 0.9347 & 10 & 0.0213 & 1.1487 & 10 \\
HARQF–SKST     & 0.0198 & 0.9142 & 11 & 0.0211 & 1.1261 & 11 \\
ARFIMA–SKST    & 0.0209 & 1.0359 & 12 & 0.0223 & 1.2275 & 12 \\
LSTM–RV–H      & 0.0147 & 0.5311 & 5 & 0.0151 & 0.7212 & 5 \\
LSTM–H         & 0.0148 & 0.5893 & 6 & 0.0153 & 0.8054 & 6 \\
HAR–H          & 0.0198 & 0.9124 & 13 & 0.0212 & 1.1214 & 13 \\
HARQ–H         & 0.0205 & 0.9817 & 14 & 0.0218 & 1.1965 & 14 \\
HARQF–H        & 0.0202 & 0.9615 & 15 & 0.0215 & 1.1728 & 15 \\
ARFIMA–H       & 0.0210 & 1.0413 & 16 & 0.0224 & 1.2349 & 16 \\
\bottomrule
\end{tabular}
\end{table}

Note: In addition to the models listed above, the paper also conducted other robustness tests. The $p$-values and test statistics suggest that the LSTM–RV–EVT model passes both the unconditional coverage and conditional coverage tests at the 1\% significance level.
From Table 3, we can conclude:  
(1) For both long and short positions, the LSTM–RV–EVT model’s empirical violation ratio is closest to the theoretical violation ratio of 0.01. Its $J_c(q)$ value is the smallest and ranks highest, indicating that it passes the backtest most successfully and outperforms the other VaR models.  
(2) Across various models and position types, the violation ratios and VaR backtesting results for long positions are generally better than those for short positions, showing that volatility on the left tail of the return distribution is typically higher than on the right tail.  
(3) Combining VaR results and volatility forecasting results, we find that ARFIMA outperforms HAR in volatility prediction. However, in extreme risk prediction, HAR still yields more accurate overall VaR predictions. ARFIMA performs better than HAR in volatility prediction but worse in VaR accuracy for extreme tail risks.  

Overall, the constructed LSTM–RV volatility prediction model and LSTM–RV–EVT VaR risk measurement model outperform traditional time series models in both volatility forecasting and VaR backtesting, highlighting the predictive advantages of deep learning models.

\section{Conclusion}

This paper studies VaR-based financial risk management from a deep learning perspective. Based on high-frequency trading data, we construct the LSTM–RV volatility prediction model and use EVT semi-parametric methods to estimate return distribution quantiles, building the LSTM–RV–EVT VaR risk measurement model. We also compare it with LSTM–EVT, HAR–EVT, HARQ–EVT, HARQF–EVT, ARFIMA–EVT, and LSTM–EVT models. The results show:  
(1) Based on trading volume information and long-memory features, the LSTM model constructed using deep learning theory has better volatility forecasting accuracy than traditional HAR, HARQ, HARQF, and ARFIMA models. The semi-parametric EVT method combined with LSTM–RV produces more accurate VaR risk measurement results than LSTM–EVT and other EVT-based models.  
(2) For VaR backtesting, in both long and short positions, the LSTM–RV–EVT model’s results are the most accurate, outperforming all other models.  
(3) In terms of volatility and VaR prediction for extreme risks, deep learning-based LSTM–RV–EVT models outperform statistical time series models, showing that deep learning theory has clear advantages in financial risk forecasting.

\end{document}